\documentclass[journal]{IEEEtran}

\usepackage{amssymb}
\usepackage[hidelinks]{hyperref}
\usepackage{graphicx}
\usepackage[percent]{overpic}
\usepackage{url}
\usepackage{subcaption}
\usepackage[noadjust]{cite}

\usepackage{amsmath} 

\def\ie{i.e.,~}
\def\eg{e.g.,~}
\def\etal{et al.}
\DeclareMathOperator*{\argmax}{\arg\!\max}
\DeclareMathOperator*{\argmin}{\arg\!\min}

\DeclareMathOperator{\accuracy}{accuracy}
\DeclareMathOperator{\sensitivity}{sensitivity}

\DeclareMathOperator{\specificity}{specificity}
\DeclareMathOperator{\TP}{TP}
\DeclareMathOperator{\FP}{FP}
\DeclareMathOperator{\TN}{TN}
\DeclareMathOperator{\FN}{FN}

\newcommand{\countDepthExperiments}{45}
\newcommand{\countDepthStudies}{22}

\usepackage[usenames,dvipsnames]{color}
\definecolor{orange}{rgb}{1,0.5,0}
 
\definecolor{gray}{rgb}{0.5,0.5,0.5}

\begin{document}

\title{Visual Diagnosis of Dermatological Disorders: Human and Machine Performance}

\markboth{Kawahara and Hamarneh - Visual Diagnosis of Dermatological Disorders: Human and Machine Performance}{}

\author{Jeremy Kawahara and Ghassan Hamarneh
\thanks{J. Kawahara and G. Hamarneh are with the School of Computing Science, Simon Fraser University, Burnaby BC V5A
1S6, Canada
\mbox{(e-mail: jkawahar@sfu.ca}; hamarneh@sfu.ca).}
}

\maketitle

\begin{abstract}
Skin conditions are a global health concern, ranking the fourth highest cause of nonfatal disease burden when measured as years lost due to disability. As diagnosing, or classifying, skin diseases can help determine effective treatment, dermatologists have extensively researched how to diagnose conditions from a patient's history and the lesion's visual appearance. Computer vision researchers are attempting to encode this diagnostic ability into machines, and several recent studies report machine level performance comparable with dermatologists.

This report reviews machine approaches to classify skin images and consider their performance when compared to human dermatologists. Following an overview of common image modalities, dermatologists' diagnostic approaches and common tasks, and publicly available datasets, we discuss approaches to machine skin lesion classification. We then review works that directly compare human and machine performance. Finally, this report addresses the limitations and sources of errors in image-based skin disease diagnosis, applicable to both machines and dermatologists in a teledermatology setting.
\end{abstract}

\section{Introduction}
Skin disorders are the most frequent reason to visit a general practitioner in studied populations~\cite{Schofield2011} and are a recognized global health burden~\cite{Hay2014}. In 2013, approximately one in four Americans saw a physician for at least one skin condition~\cite{Lim2017}. As correctly diagnosing, or classifying, skin conditions can help narrow treatment options, dermatologists have extensively researched how to classify skin conditions from a patient's history and the visual properties of skin lesions. However, skin diseases are difficult to diagnose~\cite{Sellheyer2005}, and studies suggest an unmet demand for dermatologists~\cite{Kimball2008}. To alleviate these challenges, computer vision researchers are attempting to encode this diagnostic ability into machines~\cite{Korotkov2012}, which may lead to more reproducible and accessible diagnoses in under-served communities.

The following section provides an overview of the common imaging modalities, tasks, typical diagnostic approaches used by dermatologists, and common datasets and metrics used to evaluate the performance of automated skin disease diagnosis. Section~\ref{depth:sec:machine-classify} reviews trends in machine approaches to classify skin diseases. Section~\ref{depth:sec:human-vs-machine} presents works that directly compare humans and machine skin disease classification. Finally, Section~\ref{depth:sec:discuss} discusses the performance of humans and machines.

\subsection{Non-Invasive Imaging Modalities of the Skin}

\begin{figure*}
    \centering
    \includegraphics[width=0.99\linewidth]{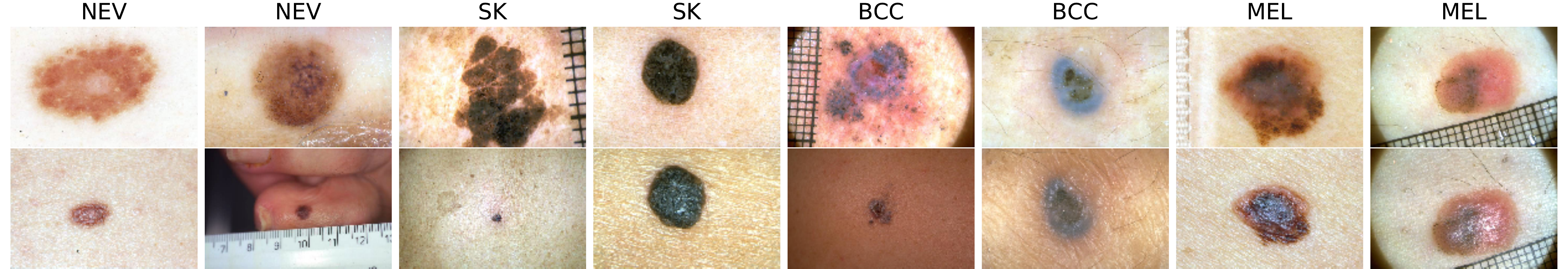}
    \caption{The same lesion (\emph{column}) can be captured as a dermoscopy (\emph{top row}) and a clinical image (\emph{bottom row}). Dermoscopy offers a more standardized acquisition, while clinical images can capture a wider field of view. These sampled images~\protect\cite{Argenziano2000} show the variability of some common lesions, where nevi (NEV) and seborrheic keratoses (SK) are benign conditions, and basal cell carcinoma (BCC) and melanoma (MEL) are common cancers.}
    \label{depth:fig:dermo-clinic}
\end{figure*}

The two common non-invasive imaging modalities to acquire skin images are clinical and dermoscopy images (Fig.~\ref{depth:fig:dermo-clinic}). \emph{Clinical} images capture what is seen with the unaided human eye and can be acquired at varying fields-of-view using non-standard cameras. \emph{Dermoscopy} (also referred to as epiluminescence microscopy~\cite{Cohen1993}) images show a magnified view of intra- and sub-epidermal structures and are acquired using a dermatoscope, which offers a more controlled field-of-view. Dermoscopy images are commonly used to help differentiate benign from malignant lesions, whereas clinical images, with their flexible field-of-view, are more commonly used to image general skin diseases. While other non-invasive imaging modalities, such as ultrasound, have been used for skin lesion diagnosis~\cite{Wassef2013}, this survey focuses on clinical and dermoscopy images. For this report, non-image information acquired from the patient is defined as the \emph{patient history}, which includes factors such as patient age, sex, lesion location, family history, and environmental factors.

\subsection{Diagnosing Skin Diseases}
\label{depth:diagnose}
Diagnosing skin diseases is complicated. There are at least 3,000 identified varieties of skin diseases~\cite{Bickers2006} with a prevalence that varies by condition. The ``gold standard'' for skin disease diagnosis is determined through a biopsy, where a portion of the affected skin specimen is extracted and sent to dermatopathologists for analysis~\cite{Sellheyer2005}. However, biopsy requires additional time and cost to extract and analyze the lesion, and may introduce potential complications to the patient. Wahie \etal~\cite{Wahie2007} reported that 29\% of patients had complications after a skin biopsy, mainly as a result of infection. Thus dermatologists may avoid biopsy in cases with well-recognized symptoms and instead rely on data collected non-invasively.

Dermatologists consider a variety of factors in their diagnoses, including patient history and the appearance (\eg morphology, colour, textures) of the affected skin region. Entire textbooks describe approaches to diagnose skin lesions (\eg \cite{Ashton2014}), where the methods are often specific to distinct types of dermatological conditions. For example, to classify skin diseases that manifest as stains on the skin, flowcharts that encode visual properties, patient history, and the lesion's location on the body can aid in the diagnosis~\cite{DaSilva2018}.

Melanoma, which accounted for 41\% of skin related deaths in the United States in 2013~\cite{Lim2017}, receives special attention due to the mortality risk. To aid less experienced clinicians in recognizing melanoma from benign lesions, rule-based diagnostic systems have been proposed, such as the ABCD rule~\cite{Nachbar1994} and the 7-point checklist~\cite{Argenziano1998}. These simplified rule-based systems produce a melanoma score based on the physician recognizing the presence of melanoma-specific morphological characteristics within the lesion.

General practitioners receive less training in dermatology than dermatologists and are often the first point of contact for skin conditions. R\"{u}bsam~\etal~\cite{Rubsam2015} found that general practitioners reported diagnosing dermatological problems using various strategies: visual recognition, testing of different treatments, and refining the diagnosis via asking additional questions. Sellheyer \etal~\cite{Sellheyer2005} reported that dermatologists correctly diagnosed roughly twice the number of cases when compared to non-dermatologists, using the histopathology diagnoses as the reference. 

\subsection{Common Non-Invasive Dermatology Tasks}
\label{depth:common-tasks}

\begin{figure}
    \begin{subfigure}[b]{0.24\textwidth}
        \centering
        \includegraphics[width=\textwidth]{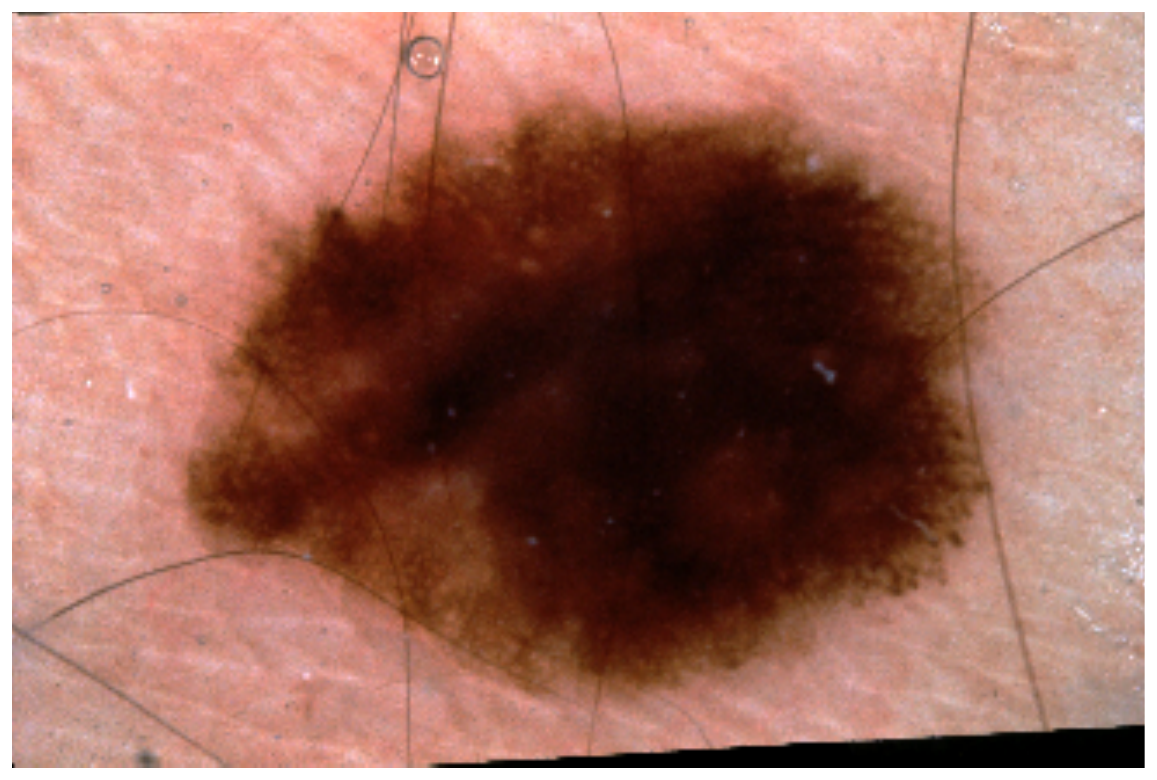}
        \caption{}
        \label{depth:fig:mel}
        \includegraphics[width=\textwidth]{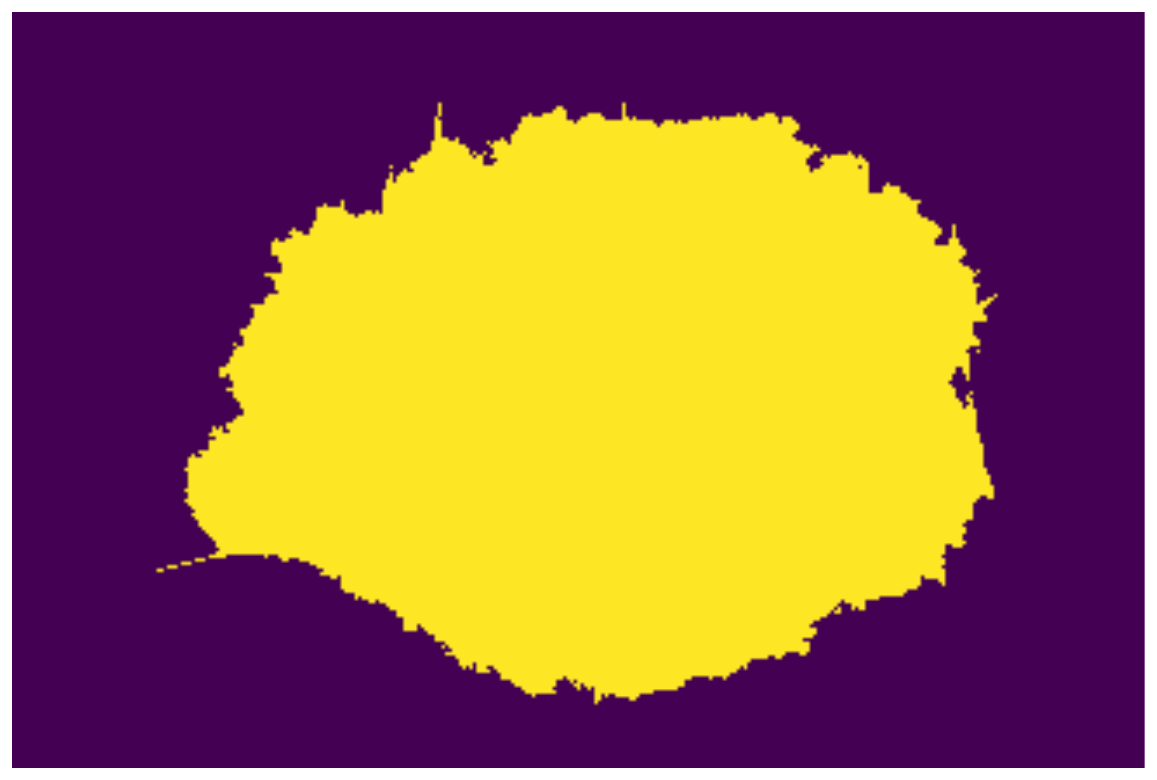}
        \caption{}
        \label{depth:fig:seg}
    \end{subfigure}
    \hfill
    \begin{subfigure}[b]{0.24\textwidth}
        \centering
        \includegraphics[width=\textwidth]{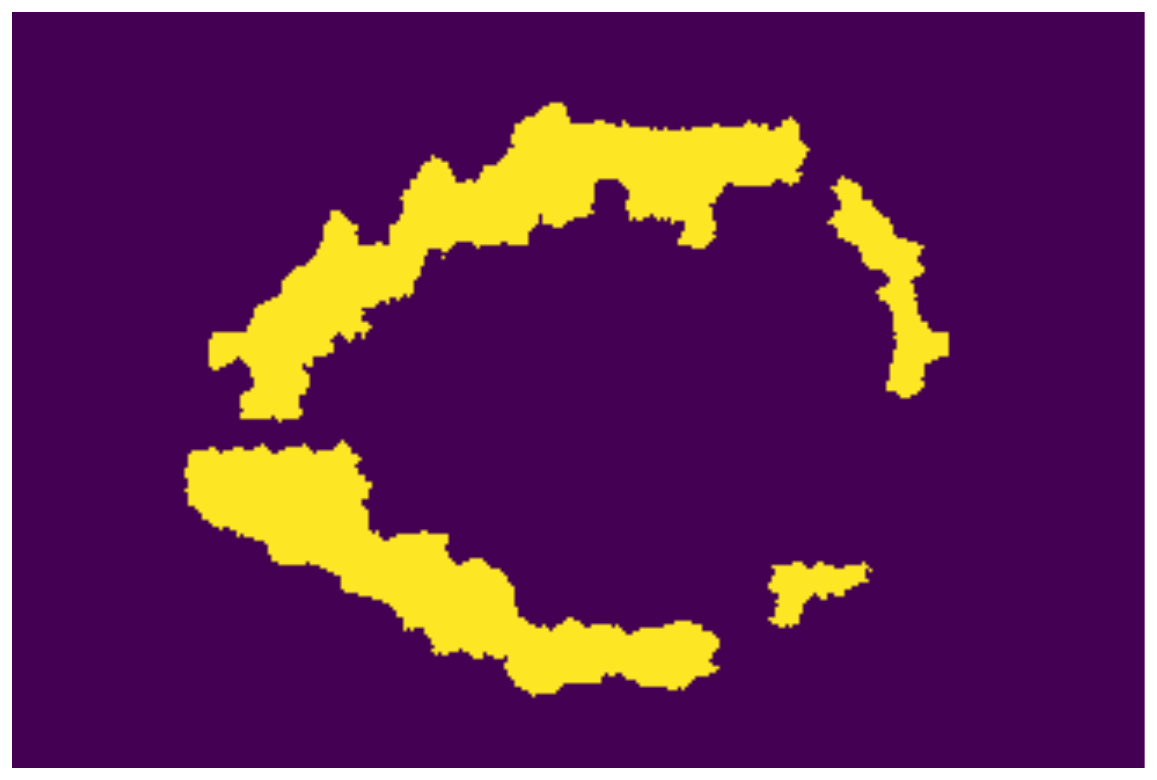}
        \caption{}
        \label{depth:fig:pignet}
        \includegraphics[width=\textwidth]{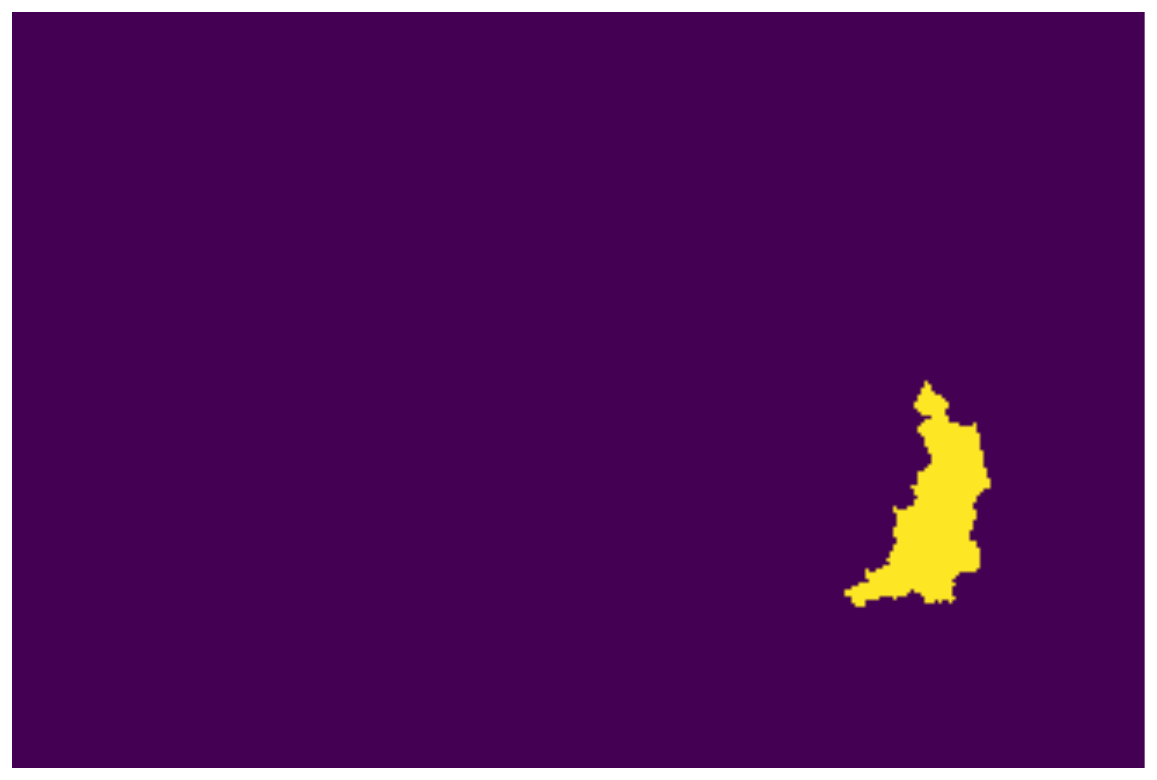}
        \caption{}
        \label{depth:fig:streaks}
    \end{subfigure}
    \caption{Common dermatological tasks: (a) \emph{classify} the observed skin lesion directly from the image; (b) \emph{segment} the lesion from the background; and, \emph{detect} the presence of dermoscopic criteria (\eg (c) pigment network and (d) streaks).}
    \label{depth:fig:common-tasks}
\end{figure}

\subsubsection{Classify Skin Diseases}
As previously discussed, to diagnose or classify a skin disease, a physician or machine predicts the type of skin disease by analyzing the patient's history, the visual properties of skin lesions, or both. Automated machine approaches to classify skin diseases is discussed in detail in Section~\ref{depth:sec:machine-classify}. While it is possible to estimate the disease class from the image directly, this process, especially during machine classification, has traditionally been broken into the following sub-tasks.

\subsubsection{Classify Dermoscopic Criteria}
\label{depth:sec:classify-dermo-criteria}
The existence of certain visual properties within a lesion may indicate a condition. For example, the presence of certain dermoscopic criteria (such as an atypical pigment network or irregular streaks) within a lesion is indicative of melanoma. Thus, one approach to classify melanoma is to classify dermoscopic criteria known to be associated with melanoma. If a lesion contains a number of these criteria, a diagnosis of melanoma can be inferred~\cite{Argenziano1998}. Approaches to classify dermoscopic criteria are discussed in Section~\ref{depth:sec:designed-features}. 

\subsubsection{Lesion Segmentation}
\label{depth:sec:localize}
Lesion segmentation,~\ie delineating the boundary of a lesion in an image (Fig.~\ref{depth:fig:seg}), allows for lesion properties to be carefully measured, and is often used to extract image features that rely on knowing the border of the lesion. Several of the works discussed in Section~\ref{depth:sec:machine-classify} segment the lesion prior to classification.

\subsubsection{Detect Dermoscopic Criteria}
A specific dermoscopic criteria (\eg streaks, which are associated with melanoma) can be both localized and classified (Fig.~\ref{depth:fig:pignet} and Fig.~\ref{depth:fig:streaks}). While this task is similar to classifying dermoscopic criteria (Sec.~\ref{depth:sec:classify-dermo-criteria}), detecting dermoscopic criteria requires localization. This task may allow physicians to localize those areas containing disease-specific criteria.

\subsubsection{Artefact Removal}
Artefact removal involves discarding potentially confounding properties from the images, and is often a preprocessing step that precedes the aforementioned tasks. For example, applying colour constancy to control for illumination~\cite{Barata2015}, and removing specular highlights~\cite{Madooei2015} or hair~\cite{Abbas2013,Mirzaalian2014} from images may improve lesion segmentation or classification.

\subsection{Common Skin Condition Image Datasets}
\label{depth:sec:datasets}
As diagnostic difficulty varies by image and type of condition, standardized datasets provide a valuable way to benchmark different approaches. Here we discuss commonly used and publicly available datasets suitable for classifying skin conditions from images. 

\begin{table*}[ht]
\centering
\caption{Details of the Atlas of Dermoscopy dataset. The two left columns show the labels for each criteria in the 7-point checklist. The right column shows the labels that correspond to the overall diagnoses. The \emph{7pt} column indicates the contribution to the 7-point checklist score, where a non-zero score indicates a criteria label associated with melanoma. The \emph{\#cases} column indicates the number of cases with the specific label.}
\setlength{\tabcolsep}{1pt}
\begin{tabular}{@{\hspace{0.33\tabcolsep}}l@{~}r@{~}r@{~}|@{~}l@{~}r@{~}r@{~}|@{~}l@{~}r@{~}}

\hline
                                              Name & 7pt &  \#cases &                                           Name & 7pt & \#cases &                                Name &  \#cases \\
\hline
       \multicolumn{3}{@{\hspace{0.33\tabcolsep}}l|@{~}}{1. \emph{Pigment Network}}          &  \multicolumn{3}{l|@{~}}{5. \emph{Vascular Structures}}            &  \multicolumn{2}{l@{~}}{\emph{Diagnosis}} \\
                                            absent &   0 &      400 &                                           absent &     0 &        823 &                  Basal Cell Carcinoma &          42 \\
                                           typical &   0 &      381 &                                       arborizing &     0 &         31 &                            Blue Nevus &          28 \\
                                          atypical &   2 &      230 &                                            comma &     0 &         23 &                           Clark Nevus &         399 \\
 \multicolumn{3}{@{\hspace{0.33\tabcolsep}}l|@{~}}{2. \emph{Regression Structures}} &                                          hairpin &     0 &         15 &                        Combined Nevus &          13 \\
                                            absent &   0 &      758 &                                within regression &     0 &         46 &                      Congenital Nevus &          17 \\
                                        blue areas &   1 &      116 &                                           wreath &     0 &          2 &                          Dermal Nevus &          33 \\
                                       white areas &   1 &       38 &                                           dotted &     2 &         53 &                       Recurrent nevus &           6 \\
                                      combinations &   1 &       99 &                                 linear irregular &     2 &         18 &                   Reed or Spitz Nevus &          79 \\
          \multicolumn{3}{@{\hspace{0.33\tabcolsep}}l|@{~}}{3. \emph{Pigmentation}} &    \multicolumn{3}{l|@{~}}{6. \emph{Dots and Globules}} &                              Melanoma &         252 \\
                                            absent &   0 &      588 &                                           absent &     0 &        229 &                        Dermatofibroma &          20 \\
                                   diffuse regular &   0 &      115 &                                          regular &     0 &        334 &                               Lentigo &          24 \\
                                 localized regular &   0 &        3 &                                        irregular &     1 &        448 &                             Melanosis &          16 \\
                                 diffuse irregular &   1 &      265 &              \multicolumn{3}{l|@{~}}{7. \emph{Streaks}} &                         Miscellaneous &           8 \\
                               localized irregular &   1 &       40 &                                           absent &     0 &        653 &                       Vascular Lesion &          29 \\
     \multicolumn{3}{@{\hspace{0.33\tabcolsep}}l|@{~}}{4. \emph{Blue Whitish Veil}}           &                                          regular &     0 &        107 &                  Seborrheic Keratosis &          45 \\
                                            absent &   0 &      816 &                                        irregular &     1 &        251 &                                       &          \\
                                           present &   2 &      195 &                                                  &       &            &  \emph{Total Cases}                   &  \emph{1011}        \\
\hline
\end{tabular}
\label{depth:tbl:dataset:7pt}
\end{table*}

\subsubsection{Atlas of Dermoscopy}
The Atlas of Dermoscopy, also know as the EDRA atlas, was originally released as a tool to instruct physicians to diagnose skin lesions and recognize dermoscopic criteria related to melanoma~\cite{Argenziano2000}. This dataset provides 1,011 cases of skin lesions, with corresponding clinical and dermoscopy images for nearly every case, patient history (\eg age, sex), and ground truth diagnoses and dermoscopic criteria labels. Table~\ref{depth:tbl:dataset:7pt} provides details on the number of cases available for each dermoscopic criteria and diagnosis. This dataset is available online~\cite{7-point-web}.

\begin{table}[ht]
\centering
\caption{The number of images for each skin disease type in the Dermofit Image Library.}
\begin{tabular}{@{~}l@{~}r|l@{~}r@{~}}
\hline
                      Name &  \# imgs &                   Name &  \# imgs \\
\hline
         Actinic Keratosis &       45 &       Malignant Melanoma &         76 \\
      Basal Cell Carcinoma &      239 &        Melanocytic Nevus &        331 \\
            Dermatofibroma &       65 &       Pyogenic Granuloma &         24 \\
               Haemangioma &       97 &    Seborrhoeic Keratosis &        257 \\
 Intraepithelial Carcinoma &       78 &  Squamous Cell Carcinoma &         88 \\
\hline
\end{tabular}
\label{depth:tbl:dermofit}
\end{table}

\subsubsection{Dermofit Image Library}
\label{depth:sec:dermofit}
The Dermofit Image Library~\cite{Ballerini2013} is available online~\cite{Edinburgh} and consists of 1,300 clinical images covering 10 classes of skin lesions (described in Table~\ref{depth:tbl:dermofit}). Images are captured in a standardized way, controlling for illumination and distance to the lesion. Manually segmented lesions are also available.

\begin{table}[ht]
\centering
\caption{The diagnosis, dermoscopic criteria, and the number of images with each label in the PH$^2$ dataset.}
\setlength\tabcolsep{0.5pt}
\begin{tabular}{@{\hspace{1\tabcolsep}}l@{\hspace{1\tabcolsep}}r@{~}|@{~}l@{\hspace{1\tabcolsep}}r@{~}|@{~}l@{\hspace{1\tabcolsep}}r@{\hspace{1\tabcolsep}}}
\hline
Name & \#imgs & Name & \#imgs & Name & \#imgs \\
\hline
         \multicolumn{2}{@{\hspace{1\tabcolsep}}l@{\hspace{1\tabcolsep}}|@{~}}{\emph{Diagnosis}}         &     \multicolumn{2}{@{\hspace{1\tabcolsep}}l@{~}|@{~}}{3. \emph{Dots/Globules}}           &  \multicolumn{2}{@{\hspace{1\tabcolsep}}l@{~}}{6. \emph{Asymmetry}}           \\
                                 Common Nevus &      80 &                                       Absent &        87 &                       Fully Symmetric &       117 \\
                               Atypical Nevus &      80 &                                     Atypical &        59 &                 Asymmetry in One Axis &        31 \\
                                     Melanoma &      40 &                                      Typical &        54 &                       Fully Asymmetry &        52 \\
   \multicolumn{2}{@{\hspace{1\tabcolsep}}l@{\hspace{1\tabcolsep}}|@{~}}{1. \emph{Pigment Network}}          &           \multicolumn{2}{@{\hspace{1\tabcolsep}}l@{~}|@{~}}{4. \emph{Streaks}}            &     \multicolumn{2}{@{\hspace{1\tabcolsep}}l@{~}}{7. \emph{Colors}}            \\
                                     Atypical &     116 &                                       Absent &       170 &                                 White &        19 \\
                                      Typical &      84 &                                      Present &        30 &                                   Red &        10 \\
 \multicolumn{2}{@{\hspace{1\tabcolsep}}l@{\hspace{1\tabcolsep}}|@{~}}{2. \emph{Blue Whitish Veil}}          &  \multicolumn{2}{@{\hspace{1\tabcolsep}}l@{~}|@{~}}{5. \emph{Regression Areas}}            &                           Light-Brown &       139 \\
                                       Absent &     164 &                                       Absent &       175 &                            Dark-Brown &       156 \\
                                      Present &      36 &                                      Present &        25 &                             Blue-Gray &        38 \\
                                              &         &                                              &           &                                 Black &        42 \\
\hline
\end{tabular}
\label{depth:tbl:ph2}
\end{table}

\subsubsection{PH$^2$}
PH$^2$ is a publicly available~\cite{Ferreira2012} dataset of 200 dermoscopy images of skin lesion. Each lesion was manually segmented and expertly labeled with a diagnosis and seven dermoscopic criteria~\cite{Mendonca2013,Mendonca2015}. These dermoscopic criteria are a subset of the 7-point checklist~\cite{Argenziano1998} and includes additional criteria relevant to other diagnostic procedures (\eg ABCD rule~\cite{Nachbar1994}). Table~\ref{depth:tbl:ph2} shows the number of images labeled with each diagnosis and dermoscopic criteria.

\begin{table}[ht]
\centering
\caption{The diagnosis labels for the HAM10000 dataset.}
\begin{tabular}{lr}
\hline
                                           Name &  \# imgs \\
\hline
 Actinic Keratosis \& Intraepithelial Carcinoma &  327 \\
 Basal Cell Carcinoma &  514 \\
 Benign Keratosis &  1099 \\
 Dermatofibroma &  115 \\
 Melanoma &  1113 \\
 Melanocytic Nevus &  6705 \\
 Vascular Lesion &  142 \\
\hline
\end{tabular}
\label{depth:tbl:ham}
\end{table}

\subsubsection{ISIC Challenge}
\label{depth:sec:isic}
The ISIC Challenge is a public dermatology competition with three tasks: segment lesions; detect dermoscopic criteria; and classify lesions (as described in Section~\ref{depth:common-tasks}). The challenge has run in 2016~\cite{Gutman2016}, 2017~\cite{Codella2018}, and 2018~\cite{Codella2019}. The dataset contains dermoscopy images, lesion segmentation masks, and dermoscopic criteria masks. Each task has standard evaluation metrics and training, validation, and testing dataset partitions. In 2018, the ISIC skin lesion classification challenge used the HAM10000 dataset~\cite{Tschandl2018} as the training set, which provides 10,015 dermoscopy images covering seven diagnosis categories. Table~\ref{depth:tbl:ham} shows the number of images with each diagnosis label in the HAM10000 dataset.

\subsubsection{SD Clinical Dataset}
The SD-198 dataset~\cite{sd198Web} consists of 6,584 clinical images covering 198 fine-grained categories of skin diseases, where each category has between 10 and 60 images~\cite{Sun2016}. Images were collected from the website DermQuest~\cite{dermquestWeb}. SD-128 is a subset of SD-198 and consists of 5,619 clinical images, where each class has at least 20 images.

\subsubsection{Others}
The \emph{Skin Cancer Detection} dataset~\cite{Waterlooskin} provides 119 melanoma and 87 non-melanoma clinical images along with lesion segmentation masks, which were gathered from two online sources: the Dermatology Information System~\cite{dermISWeb} and DermQuest~\cite{dermquestWeb}. The \emph{MED-NODE} dataset~\cite{medNodeOnline,Giotis2015} provides 70 melanoma and 100 nevi clinical images. The \emph{Melanoma Classification Benchmark}~\cite{BrinkerWeb,Brinker2019} sourced 100 dermoscopy images from the ISIC challenge~\cite{Gutman2016} and 100 clinical images from the MED-NODE dataset~\cite{medNodeOnline} such that for each type of image, 80 images are of benign nevi and 20 are of melanoma. The diagnostic performance of dermatologists over the same dataset is also provided (157 dermatologists for dermoscopy images, 145 dermatologists for clinical images). 

\subsection{Common Metrics for Classification}
Measuring the performance of a classifier on a diagnostic task is complicated as there are multiple classes of diseases, datasets are often imbalanced, and the clinical penalty for misdiagnosis may differ.

\begin{figure}
    \centering
    \begin{subfigure}[b]{0.15\textwidth}
        \centering
        \includegraphics[width=\textwidth]{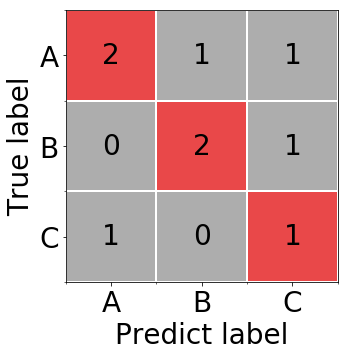}
        \caption{}
        \label{depth:fig:acc}
    \end{subfigure}
    \hfill
    \begin{subfigure}[b]{0.15\textwidth}
        \centering
        \includegraphics[width=\textwidth]{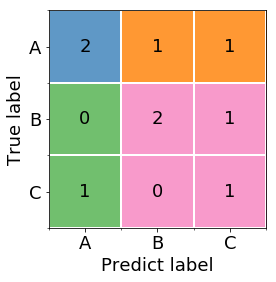}
        \caption{}
        \label{depth:fig:conf-example}
    \end{subfigure}
    \hfill
    \begin{subfigure}[b]{0.15\textwidth}
        \centering
        \includegraphics[width=\textwidth]{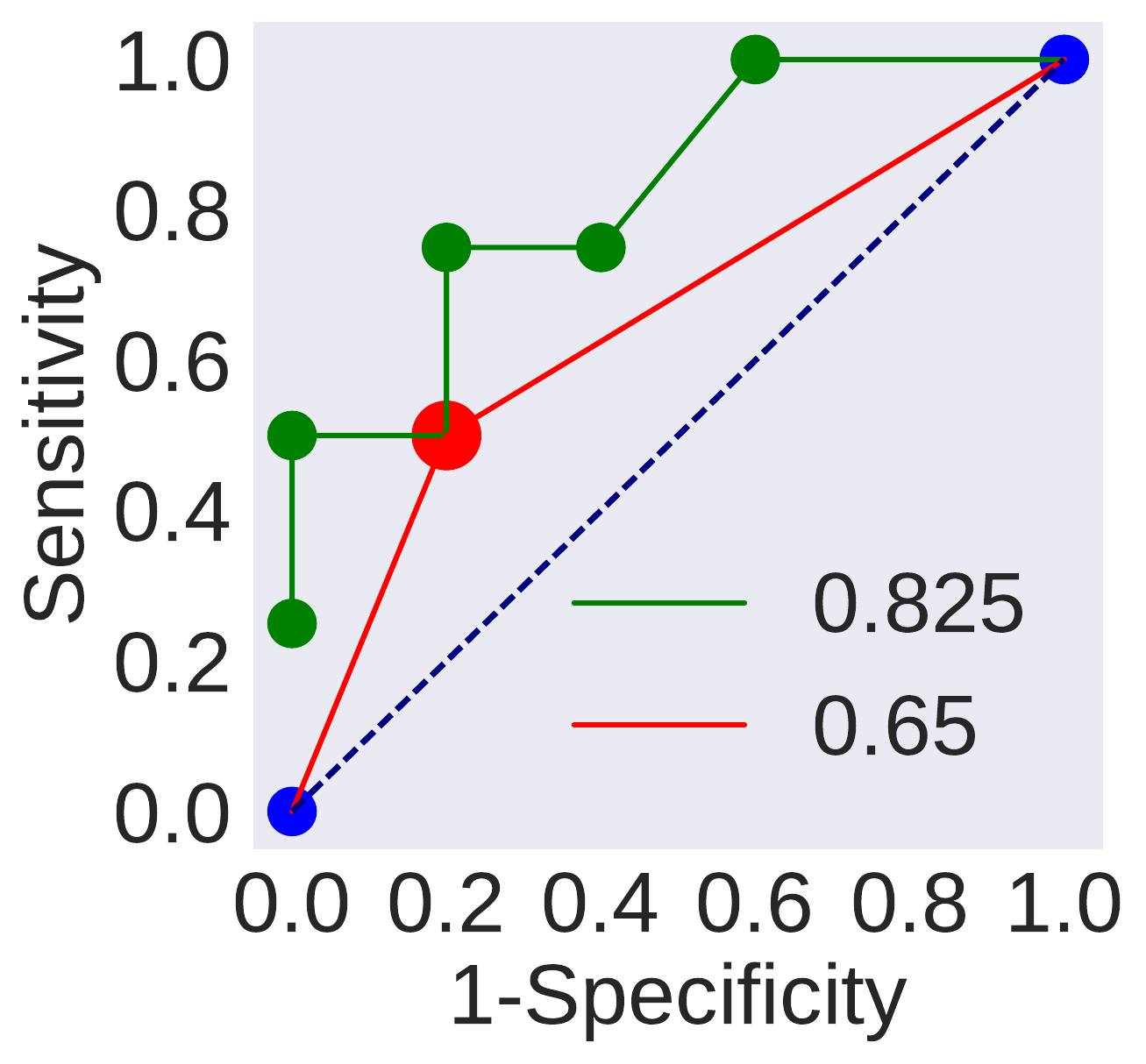}
        \caption{}
        \label{depth:fig:auroc}
    \end{subfigure}
    \caption{Computing common classification metrics. (a) Given a confusion matrix of three-classes, accuracy is computed by dividing the sum of the diagonal cells (red) with the total number of samples within all cells. (b) Considering ``A'' as a positive class, the blue cell indicates true positives, orange cells indicate false negatives, green cells indicate false positives, and pink cells indicate true negatives. (c) The AUROC curve with respect to a single class ``A''. Discrete predictions (\eg red point) may have a different ROC curve than probabilistic predictions (\emph{green line}). The values in the legend are the AUROC of the corresponding curves.}
    \label{depth:fig:metrics}
\end{figure}

Accuracy is a common metric that measures the fraction of cases where the predicted diagnoses $\hat{y}$ correctly matches the true diagnoses $y$. Some clinical studies (\eg \cite{Weingast2013}) report results that include a differential diagnosis (\ie when a physician makes more than one disease diagnosis), where the prediction is considered correct if any of the $K$ diagnoses match the true diagnosis. The top-$K$ accuracy is defined as, 
\begin{equation}
    \accuracy(\hat{y},y,K) = \frac{1}{N} \sum_{i=1}^N \sum_{k=1}^K \delta ( \hat{y}^{(i)}, y^{(i)}_{k} )
    \label{depth:eq:top-k-acc}
\end{equation}
where there are $N$ cases; $\delta(a,b)$ is the Kronecker delta function which returns 1 if $a = b$, else 0; $y^{(i)}$ is the true diagnosis for the $i$-th case; and, $\hat{y}^{(i)}_{k}$ is the $k$-th predicted differential diagnosis of the $i$-th case. Given a confusion matrix of the predicted and true labels, the top-1 accuracy can be computed by dividing the sum of the diagonal values by $N$ (Fig.~\ref{depth:fig:acc}).

Other common metrics for classification problems are sensitivity,
\begin{equation}
    \sensitivity(\hat{y},y,c) = \frac{\TP(\hat{y},y,c)}{\TP(\hat{y},y,c)+\FN(\hat{y},y,c)}
\end{equation}
and specificity, 
\begin{equation}
  \specificity(\hat{y},y,c) = \frac{\TN(\hat{y},y,c)}{\TN(\hat{y},y,c)+\FP(\hat{y},y,c)}
\end{equation}
where each metric is computed with respect to a class label $c$ (\eg a diagnosis $y$ may have $C$ possible class labels). Given $c$ as the positive class label, the number of true positives, false positives, true negatives, and false negatives are computed as,
\begin{align}
\TP(\hat{y},y,c) &= \sum_{i=1}^N \left( \delta(\hat{y}^{(i)},c) \cdot \delta(y^{(i)},c) \right)
\\
\FP(\hat{y},y,c) &= \sum_{i=1}^N \left(\delta(\hat{y}^{(i)},c) \cdot (1-\delta(y^{(i)},c)) \right)
\\
\TN(\hat{y},y,c) &= \sum_{i=1}^N \left( (1-\delta(\hat{y}^{(i)},c)) \cdot (1-\delta(y^{(i)},c)) \right)
\\
\FN(\hat{y},y,c) &= \sum_{i=1}^N \left( (1-\delta(\hat{y}^{(i)},c)) \cdot \delta(y^{(i)},c) \right)
\end{align}
respectively. Fig.~\ref{depth:fig:conf-example} shows an example using a confusion matrix. 

Another metric used to measure the performance over a public skin dataset is balanced accuracy. This metric is equivalent to the sensitivity averaged across classes,
\begin{equation}
    \overline{\sensitivity}(\hat{y},y) = \frac{1}{C} \sum_{c=1}^C \sensitivity ( \hat{y}^{(i)},y^{(i)},c )
\end{equation}
where $C$ is the number of unique classes. The averaged sensitivity assumes an equal importance for each class and may be more suitable for imbalanced datasets than accuracy (Eq.~\ref{depth:eq:top-k-acc}), as accuracy assumes an equal importance for each image.

The area under the receiver operator characteristic curve (AUROC) considers the sensitivity and specificity for a given positive class $c$ over all thresholds of the model's predicted probabilities  (Fig.~\ref{depth:fig:auroc}). The area under the resulting ROC curve is a commonly reported metric in skin lesion classification studies~\cite{Codella2018,Esteva2017,Han2018c}. As the AUROC curve considers all decision thresholds, this metric is sensitive to the predicted probabilities.

One challenge that arises when comparing the performance of humans and machines is that humans, in general, report a single discrete \emph{prediction}, while machines give a \emph{probability} distribution. Specifically, the predicted label $\hat{y}^{(i)}$ of the $i$-th lesion is defined as the most probable label within the predicted probability distribution $p^{(i)}$,
\begin{equation}
    \hat{y}^{(i)} = j^* = \argmax_{j \in \{1, \dots, C\}} p^{(i)}_{j}
    \label{depth:eq:predictions}
\end{equation}
where $C$ is the number of classes, and $p^{(i)}_{j}$ is the $j$-th class probability of the $i$-th lesion. 

When computing the AUROC curve for a human, the sensitivity and specificity of the predictions are used and the ROC curve is assumed to be linear (see Fig.~\ref{depth:fig:auroc}). In contrast, the probabilistic outputs of machines often produce non-linear ROC curves (\eg \cite{Esteva2017}). In addition, while the ROC curve demonstrates the \emph{limits} of the model's ability to discriminate~\cite{Zweig1993}, this considers all possible probability thresholds, rather than the actual predictions. 
Thus a probabilistic model that makes incorrect predictions can still achieve a high AUROC score. Further, in a multi-class scenario where the non-positive classes are all considered negative, a ROC curve may be influenced by class imbalances~\cite{Fawcett2006}.

While other metrics, such as average precision, are used~\cite{Demyanov2017,Gutman2016}, they are reported less frequently in the literature. In order to compare human and machine predictions over multi-class datasets across a variety of works, we focus on reviewing experiments within studies where diagnostic accuracy can be inferred. Limitations when relying on diagnostic accuracy are discussed in Section~\ref{depth:sec:metric-limits}.

\section{Machine Skin Disease Classification}
\label{depth:sec:machine-classify}
This section primarily focuses on skin disease classification and discuss other tasks (\eg segmentation) in the context of supporting classification. A classification system is seen as a pipeline or model $\phi$ and parameters $\theta$ of $\phi$, and generally requires a dataset of the observable input data $x$ (\eg images, patient history), and, for training or evaluation, the desired output data $y$ (\eg disease diagnosis). To design a classification system, a general optimization is done,
\begin{equation}
    \phi^*, \theta^* = \argmin_{\phi, \theta} E(\phi(x; \theta), y)
    \label{depth:eq:general}
\end{equation}
where $\phi(\cdot)$ is a model or pipeline that transforms the input data $x$ into a predicted output $\hat{y} = \phi(\cdot)$, such that $\hat{y}$ matches the desired output $y$. $\theta$ are the parameters for the model/pipeline; $E(\cdot)$ measures the error between the predicted $\hat{y}$ and true output $y$, and can encode prior knowledge about the output or model parameters (\eg regularization). $\phi^*, \theta^*$ are the found model/pipeline and parameters, respectively, that minimize (globally or locally) $E(\cdot)$. There are many ways to optimize Eq.~\ref{depth:eq:general} as can be found in recent surveys~\cite{Barata2018,Brinker2018,Celebi2019,Mishra2016,Okur2018,Pathan2018,Sultana2018}. A common approach is for a human to design a fixed model/pipeline $\phi$, and to learn the parameters $\theta$ from the data using an explicit optimization (\eg gradient descent). In the following sections, we discuss common pipelines/models, parameters, and optimization approaches, where each proposed component can be thought to be part of the general optimization of Eq.~\ref{depth:eq:general}.

\subsection{Sequential Pipeline Approach}
Celebi \etal~\cite{Celebi2007} proposed the following general pipeline $\phi$ to classify dermoscopy images as either benign or melanoma: 1) segment the skin lesion; 2) extract colour and shape-based features from the lesion border and regions within the lesion; 3) select a subset of discriminative features; and 4) use a machine learning classifier to distinguish among the classes. Related to Eq.~\ref{depth:eq:general}, the pipeline and design choices $\phi$ are chosen by the authors (\eg types of colour features to extract~\cite{Madooei2016}), while the learned parameters $\theta$ are explicitly optimized by a support vector machine classifier (SVM). 

\begin{figure}
    \centering
    \includegraphics[width=0.98\linewidth]{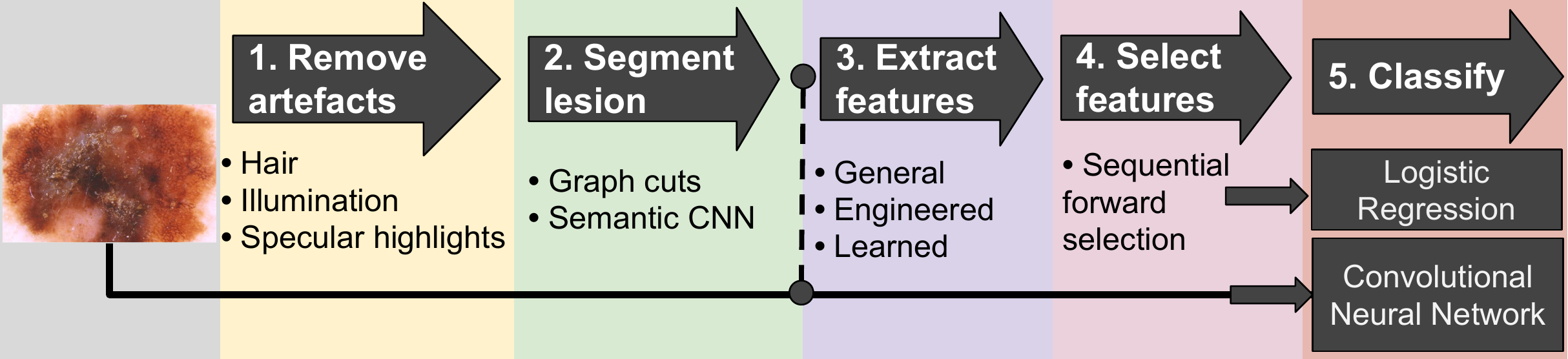}
    \caption{A common general pipeline to classify skin diseases. Image features can be extracted, then sent to a classifier (\eg logistic regression). CNNs can extract features and classify diseases directly from images, but may also be used to extract features.}
    \label{depth:fig:skin-flow}
\end{figure}

This is referred to as a \emph{sequential pipeline} approach since it follows a series of well-defined steps, where the fixed output from one step becomes the input to another (Fig.~\ref{depth:fig:skin-flow}). 

Many existing works propose a variation on this sequential pipeline, where a step may be improved or omitted. For example, Ballerini~\etal~\cite{Ballerini2012,Ballerini2013} used 960 clinical images from the Dermofit dataset (Sec.~\ref{depth:sec:dermofit}) to classify among five types of skin lesions with a 74.3\% accuracy. They segmented lesions using a region-based active contour approach, extracted human engineered colour and texture features from the lesion and healthy skin separately, and selected features using sequential forward feature selection~\cite{Jain1997}. A hierarchical $k$ nearest neighbour classifier clusters the images into two high-level classes (benign vs. pre-malignant and cancer), followed by another classification to determine the sub-classes. Leo~\etal~\cite{Leo2015} evaluated over 1,300 images of Dermofit composed of 10-classes, and followed a similar approach to achieve a classification accuracy of 67\%. Shimizu \etal~\cite{Shimizu2015} segmented lesions and removed artefacts using a colour thresholding based approach. They extracted 828 colour and texture features based on the sub-regions of the segmented lesions, applied feature selection to select a subset of discriminative features, and used a two-stage hierarchical linear classifier to classify among four conditions.

\subsection{Features Designed for Specific Dermoscopic Criteria}
\label{depth:sec:designed-features}
Instead of general colour and texture-based image features, some works specifically design features to capture known salient properties of a skin condition. This is common in melanoma classification, where the presence of specific dermoscopic criteria suggests melanoma (Section \ref{depth:sec:classify-dermo-criteria}). For example, in order to detect blue-white veils in dermoscopy images, Madooei \etal~\cite{Madooei2013} matched the lesion colours to a template of common blue-white veil colours. To detect and classify the types of streaks within dermoscopy images, Mirzaalian \etal~\cite{Mirzaalian2012a} used a filter designed to capture the tubular properties of streaks. They segmented lesions using graph cuts~\cite{Boykov2006}, and used features derived from the detected streaks to train an SVM to distinguish the type of streaks. Fabbrocini \etal~\cite{Fabbrocini2014} designed separate pipelines and engineered features to classify seven dermoscopic criteria. For example, to classify irregular streaks, they segmented the lesion and compared the irregularity at the border to a reference threshold. 

\subsection{Learned Features of Dermatological Images}
\label{depth:sec:learned-features}
Rather than general engineered or features designed to target specific dermatological criteria, features can be learned from the data. In order to classify melanoma from non-melanoma in dermoscopy images, Codella \etal~\cite{Codella2015} applied an unsupervised sparse coding approach~\cite{Mairal2014} to learn a sparse number of patterns that minimized an image reconstruction error. They also passed dermoscopy images into a CNN that was trained over the natural images (\eg cats and dogs) of ImageNet~\cite{Russakovsky2014}, and extracted the CNN responses from select layers to use as feature vectors. They found that using these learned features to train an SVM gave a similar level of classification performance when compared to the previous state-of-the-art approach of using an ensemble of general engineered features. Over the clinical images of Dermofit (Sec.~\ref{depth:sec:dermofit}), Kawahara \etal~\cite{Kawahara2016} found that training a logistic regression classifier on features extracted from a pretrained CNN outperformed previously published approaches that relied on the classical pipeline and general engineered features.

Learning features directly from the images can also simplify the overall pipeline (Fig.~\ref{depth:fig:skin-flow}) as this approach does not rely on engineered image features that require careful lesion segmentation (\eg computing features at the border of the lesion). Avoiding lesion segmentation may be desirable as segmentation is challenging~\cite{Celebi2015}. For example, 16\% of the lesions segmented by the top performing lesion segmentation method of the 2017 ISIC challenge had a Jaccard Index of less than 0.6, which is considered a failure~\cite{Codella2018}. These segmentation errors may propagate to errors in the features, which may decrease classification performance.

\subsection{Joint Optimization to Learn Features and Classify}
\label{depth:sec:joint}
The approaches described so far extract fixed features from the images $f(x)$, and perform a separate optimization for classification,
\begin{equation}
    \theta^* = \argmin_{\theta} E \left( \phi (f(x); \theta ), y \right)
    \label{depth:eq:classify}
\end{equation}
where $\phi(\cdot)$ is a user specified classifier (\eg SVM, logistic regression classifier) parameterized by $\theta$. The parameters learned when optimizing Eq.~\ref{depth:eq:classify} are based on the fixed (possibly learned) set of image features $f(x)$, under the assumption that they will prove useful for classification. This section looks at works that combine feature learning and classification in a single optimization.

\emph{Deep learning}~\cite{LeCun2015a} involves training a model composed of stacked layers of trainable parameters that learn non-linear feature representations of the data. Deep learning is widely used in skin lesion analysis, with the organizers of the 2017 ISIC skin challenge~\cite{Codella2018} (Section ~\ref{depth:sec:isic}) noting that among the entries of this public challenge:
\begin{quote}
``All top submissions implemented various ensembles of deep learning networks.''
\end{quote}

One type of deep learning model well suited for image classification is the CNN. The structure of the CNN considers the properties of images (locality of features, spatial invariance) and learns to transform the image pixels into discriminative feature representations. As all parameters within the CNN are learned, a CNN can be thought of as ``synthesizing their own feature extractor''~\cite{LeCun1998a}.

In this approach, a human designed CNN architecture $\phi(\cdot)$ is chosen, and an explicit optimization algorithm finds the CNN's parameters, 
\begin{equation}
    \theta^* = \argmin_{\theta} E \left(\phi(x; \theta), y \right) .
    \label{depth:eq:cnn}
\end{equation}
In contrast to Eq.~\ref{depth:eq:classify}, this equation does not have the human chosen representation of the features (\ie $f(\cdot)$ in Eq.~\ref{depth:eq:classify}). Rather, the parameters to compute the features, and the parameters to classify are learned within a single optimization.

There are many possible error functions $E(\cdot)$, but a common choice for classification (others discussed in Sec.~\ref{depth:sec:other-losses}) is the categorical cross-entropy loss function, 
\begin{equation}
    E(p,y) = - \frac{1}{N} \sum_{i=1}^N \sum_{j=1}^C y^{(i)}_{j} \mathrm{log}(p^{(i)}_{j})
    \label{depth:eq:cross-entropy}
\end{equation}
where $N$ is the number of images, $C$ is the number of classes (\eg types of skin diseases), $y^{(i)}_{j}$ is a one-hot encoded ground truth label, and $p^{(i)}_{j}$ is the predicted $j$-th class probability for the $i$-th image. Stochastic gradient descent can be used to learn the parameters $\theta$ that minimize Eq.~\ref{depth:eq:cnn}, where the parameters $\theta$ of the model $\phi$ are iteratively updated~\cite{LeCun1998a}. 

\subsubsection{CNNs for Classifying Skin Diseases}
\label{depth:sec:cnn-skin}
Many works (\eg \cite{Kawahara2016a,Menegola2017,Romero-Lopez2017}) that use CNNs to classify skin diseases rely on CNN architectures (\eg VGG16~\cite{Simonyan2015}) that perform well when classifying natural images (\eg ImageNet~\cite{Russakovsky2014}). The parameters of the CNN learned over the natural images are stored (referred to as a \emph{pretrained} CNN) and are used to initialize the weights of the CNN before training on a different target domain, such as skin images. The process of refining the learned parameters to a new target domain is referred to as \emph{transfer-learning} or \emph{fine-tuning} the CNN. 

While several CNN approaches ignore lesion segmentations~\cite{Kawahara2016a,Menegola2017,Romero-Lopez2017}, which simplifies the overall pipeline (Sec.~\ref{depth:sec:learned-features}), explicitly localizing the skin lesion prior to training a CNN may reduce distracting background artefacts and improve overall performance. Yoshida \etal~\cite{Yoshida2016} trained a CNN to classify melanoma from nevi using dermoscopy images, where the major axis of each lesion was aligned in order to better capture the lesion asymmetry that is commonly associated with melanoma. A CNN trained using image augmentations that were constrained to maintain this alignment outperformed a CNN trained on non-aligned lesions when the same amount of image augmentation was performed. Yu \etal~\cite{Yu2017} used a two-step process where the lesion is first segmented using a fully-convolutional neural network trained to segment skin lesions, then the lesion is cropped based on this segmentation and passed to a CNN for classification. Using this approach, Yu \etal~\cite{Yu2017} ranked first place on the ISBI-ISIC 2016 skin lesion classification challenge (Sec.~\ref{depth:sec:isic}). 

Although deep neural networks are often used to classify skin lesions~\cite{Codella2018}, not all groups report better performance when compared to using a separate feature extraction and classification approach (Eq.~\ref{depth:eq:classify}). Sun~\etal~\cite{Sun2016} collected 6,584 clinical and dermoscopy skin images, spanning 198 classes of common skin diseases from an online source. They trained an SVM on general engineered features and achieved a classification accuracy of 52.19\% over the 198 classes, outperforming the 50.27\% accuracy achieved using a CNN (VGG16~\cite{Simonyan2015}) pretrained over ImageNet and fine-tuned to classify the skin conditions. A similar result was also found by Yang~\etal~\cite{Yang2018}. 

Ge~\etal~\cite{Ge2017} represented skin images as concatenated l2-normalized responses from ResNet-50~\cite{He2016} and VGG16~\cite{Simonyan2015} fine-tuned on skin images. They extracted features using compact bilinear pooling~\cite{Gao2015}, and trained an SVM to classify among 15 types of skin diseases using skin lesions acquired as both a clinical and dermoscopy image, outperforming a single fine-tuned CNN. They summed the predicted probabilities from each imaging modality together to form a final prediction accuracy of 71\%. They used 24,182 training images and 8,012 testing images from an internal dataset known as ``MoleMap''.

\subsubsection{Other Classification Loss Functions}
\label{depth:sec:other-losses}
Cross-entropy is a common loss function used to train a CNN (Eq.~\ref{depth:eq:cross-entropy}); however, 
other losses are also used. Ge \etal~\cite{Ge2017a} incorporated clinical and dermoscopy images into a single CNN model trained to minimize the mean squared loss between the predicted and true vectors, and reported only minor differences in overall performance when compared to the cross-entropy loss. They used class activation maps~\cite{Zhou2016} to find salient areas of the image, and extracted dense features from the diseased area using bilinear pooling~\cite{Gao2015}. They achieved a classification accuracy of 70\% accuracy over 15-classes. 

Demyanov~\etal~\cite{Demyanov2017} trained a CNN using a tree-loss function that incorporated a human defined skin disease taxonomy. This taxonomy allows data to be labeled with different granularities. For example, a lesion could have the general label of ``benign'' (coarse granularity) and the more specific label of ``blue nevus'' (fine granularity). Using an internal dataset of 40,173 dermoscopy and clinical images, composed of 15 skin conditions, they trained ResNet-50~\cite{He2016} using their tree-loss function and obtained 64.8\% accuracy, demonstrating a small but consistent improvement to accuracy when compared to training without the tree-loss function.

\subsection{Incorporating Non-Visual Information}
While our focus in this survey is on visual classification, other non-visual information may provide important context when classifying skin diseases. Razeghi~\etal~\cite{Razeghi2014} collected answers that humans gave to 37 simple questions about skin images (\eg is the patient an infant, child, or adult?), as well as extracted general engineered features from the images. Using 2,309 clinical images from an online source composed of 44 disease types, they manually placed a bounding box around the lesion in the image, and trained a random forest to classify the skin diseases. Using only visual information, a trained random forest yielded 15.76\% accuracy. Using only the human given answers to questions yielded 16.58\% accuracy, and combining both yielded 25.12\% accuracy. 
Kawahara~\etal~\cite{Kawahara2018} incorporated clinical images, dermoscopy images, and patient meta-data (\eg lesion location, sex) in a single CNN model designed to jointly classify multiple types of dermoscopic clinical criteria (\eg type of streaks) and skin disease diagnoses. This approach reached an average classification accuracy of 73.7\% when classifying skin diseases and dermoscopic criteria, which was an improvement over training on a single modality.

\subsection{Image Retrieval}
Adopting machine diagnoses into clinical practice may be hindered if the model does not offer an intuition into how the diagnoses are made. One approach towards more interpretable models is to retrieve images of known diseases that are visually similar to a user's lesion, allowing a user to visually inspect similar images of known diseases and infer a diagnosis.

Given a test query image $q$, the goal of image retrieval is to find the image $x^{(i)}$ within a dataset of known skin diseases that is most similar to the query image $q$. The corresponding known label $y^{(i)}$ is used as the prediction  $\hat{y}^{(i)}$ for the unknown query image,
\begin{equation}
    x^{(i)}, y^{(i)} = \argmin_{i \in \{1,\dots,N\}} D(f(q),f(x^{(i)}))
\end{equation}
where $N$ is the number of samples in the labelled dataset, $f(x^{(i)})$ computes the features for the $i$-th image of the labelled dataset, and $D(a,b)$ measures the dissimilarity (\eg cosine distance) between two feature vectors. A variation on this approach is to find the $k > 1$ most similar images and a corresponding ranked list of diagnoses.

Celebi \etal~\cite{Celebi2004} retrieved similar skin images using shape features extracted from segmented lesions and weighted these features to match the human perception of similar lesion shapes. Ballerini \etal~\cite{Ballerini2010} extracted colour and texture features from skin lesions, selected and combined features using a genetic algorithm, and retrieved up to $k=10$ labelled images that had the lowest distance $D(\cdot)$ (\eg euclidean distance) in feature space $f(\cdot)$ to a given query image. 
Bunte \etal~\cite{Bunte2011} proposed an image retrieval system that retrieved dermoscopy images of similarly coloured lesions. They extracted colour-based features from manually selected patches within the lesion and healthy skin, learned features based on four classes of colours, and retrieved images using a $k$-nearest neighbourhood, where $k$ ranged from 1 to 25.

Kawahara \etal~\cite{Kawahara2017-grail} used a minimal path approach to find a progression of visually similar images between two query images. This may be useful in finding images related to disease progression (\eg from benign to malignant). Skin lesions were represented as nodes in a graph, with edges representing the visual dissimilarity between lesions in a feature space based on the responses of a pretrained CNN.

Kawahara \etal~\cite{Kawahara2018} fine-tuned a pretrained CNN to classify both the disease and the 7-point checklist criteria~\cite{Argenziano1998}. They used the CNN layer responses $f(\cdot)$ to represent images and retrieved the class from the image with the lowest cosine distance $D(\cdot)$ to a query image. Over five classes of skin diseases, they reported an averaged retrieval accuracy of 71.1\%. Tschandl \etal~\cite{Tschandl2018a} used a similar approach for dermoscopy images and found that image retrieval had comparable accuracy with classification and allowed for better recognition of diseases that occur in datasets that the CNN was not trained on.

\section{Dermatologist and Machine Performance}
\label{depth:sec:human-vs-machine}
This section examines works that report the skin lesion classification performance of human dermatologists and machines from the same dataset. Dermatology studies that report lesion diagnosis via static images are also reported. 

\subsection{Ground Truth for Dermatologists' Classifications}
Studies that measure human performance have dermatologists make diagnoses based on the provided static information (\eg images, curated patient history). These diagnoses are compared to the ``ground truth'' class labels, which are determined by more rigorous diagnoses procedures. These procedures vary, but often consists of diagnoses by histopathology, a consensus of experts, interactive face-to-face sessions between dermatologists and patients, or a combination of approaches~\cite{Tschandl2018}.

\subsection{Dermatologists Compared with Machine}
\label{depth:sec:derm-vs-machine}
Ferris \etal~\cite{Ferris2015} manually segmented skin lesions, extracted shape, colour, and texture based features, and trained a decision forest over 273 dermoscopy images, and tested the classification performance on 40 benign and 25 malignant dermoscopy images. Over the same test dataset of 65 lesions, 30 clinicians had an averaged melanoma sensitivity of 70.8\% and specificity of 58.7\%, whereas the automatic classifier had a melanoma sensitivity of 96\% and specificity of 42.5\%.

Codella~\etal~\cite{Codella2017} and Marchetti \etal~\cite{Marchetti2017} compared an ensemble of top performing machine classification approaches, which included CNNs, to the average of eight dermatologists. 
Over 100 dermoscopy test images, the automated system achieved a higher accuracy (76\%) than dermatologists (70.5\%) when classifying 50 melanoma from 50 benign neoplasm images~\cite{Codella2017}. The eight dermatologists achieved an averaged sensitivity of 82\% and specificity of 59\%, while five top performing automated approaches on the ISIC-2016 challenges achieved a voting average sensitivity of 56\% and specificity of 90\%. 

Esteva~\etal~\cite{Esteva2017} collected a dataset of 129,450 clinical images, which included 3,374 dermoscopy images, and spanned across 2,032 types of skin diseases. They grouped classes together based on their clinical similarity as per a human-defined taxonomy, which yielded 757 partitions (classes) for training. They used the Inception-V3 architecture~\cite{Szegedy2016}, pretrained over ImageNet~\cite{Russakovsky2014}, and fine-tuned the model on the partitioned classes. They reported results over different levels of the taxonomy, where the sum of the predicted probabilities in the descendant nodes determined the higher level classes predicted probabilities. Over a nine-class partition of dermatologist inspected images (\ie may not be verified via histopathology), the CNN achieved an overall skin disease classification accuracy of 55.4\%, which is comparable to the accuracy achieved by two dermatologists (53.3\% and 55.0\%).

To compare a CNN with humans in similar training conditions, Tschandl~\etal~\cite{Tschandl2017} showed 298 dermoscopy images from six different lesion classes to 27 medical students (without prior dermoscopy experience), and provided only the corresponding diagnosis of each image without explaining the diagnostic features. The same images were used to fine-tune an Inception-V3 CNN architecture~\cite{Szegedy2016} pretrained over ImageNet, where the last layer was replaced to match the target classes. Both the students and CNN then diagnosed the diseases from 50 test images. The CNN achieved a diagnostic accuracy of 69\% and was reported to demonstrate a similar diagnostic agreement as the average agreement among students. When diagnosing malignant lesions (basal cell carcinoma and melanoma) from benign, the CNN achieved a higher sensitivity (90\% for CNN, 85\% for students), but lower specificity (71\% for CNN, 79\% students) than the students' average scores. 

Han \etal~\cite{Han2018} formed a dataset of 49,567 hand and foot nail images by using manually labelled data, assisted by first training a hand and foot CNN classifier, followed by a region-CNN~\cite{Ren2016} trained to localize the nail plate, and an image quality CNN that eliminates poor quality nail images. They showed that a CNN could classify nail images that contain onychomycosis (a nail fungal infection) from other nail disorders with a higher Youden Index (sensitivity + specificity - 1) (67.62\%) than then the average of 42 human dermatologists (48.39\%) over 1133 images.

Han \etal~\cite{Han2018c} fine-tuned a pretrained CNN (ResNet-152~\cite{He2016}) on 19,389 manually cropped clinical images taken from primarily an Asian population (Asan dataset). The training dataset was composed of 248 classes of skin diseases, while testing was done on an aggregated 12-class subset. The CNN tested over images from an Asian population achieved an accuracy of 57.3\%, and 55.7\% over the 12-classes of the Asan dataset, and the 10-classes of Dermofit (Sec.~\ref{depth:sec:dermofit}), respectively. Additional experiments comparing the diagnoses of 16 dermatologists over a subset of these images had, in general, a ROC curve inside the ROC curve produced by the CNN.

Yang~\etal~\cite{Yang2018} had general doctors, junior dermatologists, and expert dermatologists classify skin images from 198 classes of skin diseases. Two doctors from each category were invited to independently classify images and discuss the diagnosis when they differed. The accuracy was 49.00\% for general doctors, 52.00\% for junior dermatologists, and 83.29\% for expert dermatologists. The accuracy of the top performing CNN was 53.35\%, which was lower than the expert dermatologists, but comparable with general doctors and junior dermatologists. 

Haenssle~\etal~\cite{Haenssle2018} trained a CNN to classify dermoscopy images as either a benign nevi or melanoma using training images from a variety of sources. Using 100 test dermoscopy images, they compared the classification results of the CNN with 58 dermatologists. On average, dermatologists had a sensitivity of 86.6\% and specificity of 71.3\%, while a CNN tested over the same images achieved a sensitivity of 95\% and specificity of 63.8\%.

Brinker~\etal~\cite{Brinker2019a} compared the performance of 157 dermatologists with a CNN trained to classify melanoma within dermoscopy images. Using 100 dermoscopy images, the dermatologists had an averaged sensitivity to melanoma of 74.1\% and averaged specificity of 60.0\%. When the specificity of the CNN was set to 74.1\% the CNN achieved a specificity of 86.5\%.

Fujisawa~\etal~\cite{Fujisawa2019} fine-tuned a pretrained CNN on clinical images to classify among 21 disease classes and aggregated the predicted classes within a skin tree hierarchy. Using the diagnoses aggregated at the third level of the tree with 14 classes, the CNN achieved an accuracy of 76.5\%, outperforming the averaged diagnostic accuracy of 13-board certified dermatologists (59.7\%) and nine dermatology trainees (41.7\%).

Tschandl~\etal~\cite{Tschandl2019} combined the predictions from a CNN trained on dermoscopy and a CNN trained on clinical close-up images to form a final diagnosis. When compared with 95 human examiners with varying levels of expertise~\cite{Sinz2017}, the CNN had a higher number of correct specific diagnosis (37.6\%) than the human examiners (33.5\%), but lower than human expert dermatologists (40.0\%). 

\subsection{Comparing Dermatologists on Static Images}
\label{depth:sec:derm-static}
To better estimate human performance, this section primarily examines \emph{store-and-forward} teledermatology studies, where the patient data (\eg lesion image, patient history) is sent to a dermatologist for review~\cite{Trettel2018}, 

To compare how different types of static images influences human performance, Sinz \etal~\cite{Sinz2017} had 95 human examiners (including 62 dermatologists) classify 50 images randomly sampled from 2,072 cases into one of 51 possible diagnoses. Using clinical images, the averaged accuracy was 26.4\%. Using dermoscopy images, the averaged accuracy improved to 33.1\%, indicating that performance depends on the imaging modality. 

To compare \emph{in vivo} diagnosis and diagnosis via static images, Carli \etal~\cite{Carli2002} collected 256 lesions composed of seven classes of biopsy verified diseases. Using the consensus of two dermatologists (in disagreement, a third dermatologist was consulted), they reported a diagnosis accuracy of 40.1\% during clinical examinations without dermoscopy. When \emph{in vivo} dermoscopy was incorporated with the clinical examination, the accuracy improved to 72.3\%. The accuracy dropped to 54.7\% when the dermatologists had access to only the dermoscopy photographs and patient history, but not clinical information.

Weingast \etal~\cite{Weingast2013} had 263 patients photograph their own lesions, when possible, using a mobile camera, and provide additional questionnaire information. They collected a wide variety of skin conditions, which were typical of the authors outpatient unit. Multiple teledermatologists reviewed each case, and overall, 49\% of the gathered cases could be correctly diagnosed via teledermatology when compared to a face-to-face consultation (a differential diagnosis was allowed in some cases \ie top-2 accuracy Eq.~\ref{depth:eq:top-k-acc}). The teledermatologists reported only 61\% of the cases contained sufficient information to make a diagnosis.

In a prospective study with 63 dermoscopy images, Walker~\etal~\cite{Walker2019} used a CNN to extract visual feature representations that were converted into sound and visually or audibly analyzed by humans to detect cancerous skin lesions, achieving a sensitivity of 86\% and specificity of 91\%.

Brinker~\etal~\cite{Brinker2019} had 157 dermatologist assess 100 dermoscopy images and 145 dermatologists assess 100 clinical images composed of nevi and melanoma skin lesions. Dermatologists provided a management decision (biopsy vs. reassure patient), achieving an average of 74.1\% sensitivity and 60.0\% specificity for dermoscopy images; and, 89.4\% sensitivity and 64.4\% specificity for clinical images.

\section{Discussions}
\label{depth:sec:discuss}
This section lists the challenges of comparing across studies, summarizes the reported performance of selected human and machine skin disease classification works, and discusses potential limitations and sources of error within image-based diagnoses of skin conditions. 

\subsection{Challenges of Metrics and Comparing Skin Studies}
\label{depth:sec:metric-limits}
This section primarily focused on the metric of diagnostic accuracy as it is commonly reported (or can be inferred) in both clinical and computer vision studies, and it gives us a single intuitive metric for multi-class problems. However, relying on diagnostic accuracy assumes that all errors are equal, which may hide a poor performance on infrequently occurring diseases. Other metrics, such as averaged diagnostic sensitivity or precision, address the class imbalance problem by giving an equal weighting to each class, resulting in a higher weighting of infrequent conditions. All these metrics are limited since clinically, some conditions are more important to correctly diagnose than others (\eg a false-negative melanoma diagnosis can be fatal). One potential solution is to weight each misdiagnosis to account for the severity of a misdiagnosis. However, establishing such a clinical weighting is non-trivial for multi-class problems, and would require significant expert knowledge.

Another approach is to ignore diagnostic performance and instead focus on predicting appropriate treatments (\eg management strategies~\cite{Sinz2017}). While this considers the clinical implications of a disease, it requires a consensus on appropriate treatments, which may change as new treatments become available. Another complication is illustrated in the case of melanoma, where images that are biopsy verified are, by definition, ones that a dermatologist recommended for biopsy. Thus, biopsy images labeled as benign are clinically suspicious enough that an expert flagged them for biopsy. One may question if the goal of machine classification should be to replicate the dermatologist's decision or to classify the underlying disease.

A limitation in comparing across studies is that the difficulty of diagnosing diseases depends on the dataset (\eg some diseases display more consistent morphology), making it unclear if one particular methodology performs better of if the differences are due to the datasets.
Studies that compare the performance of both humans and machine (Sec.~\ref{depth:sec:derm-vs-machine}) often compare over the same dataset, allowing for a fairer comparison.

Nevertheless, with these limitations stated, diagnostic accuracy is used as our primary metric, largely due to insufficient information provided in many studies to infer other metrics and the challenges associated with choosing a single more descriptive metric. Results are aggregated across different studies, composed of a variety of datasets, in order to compare the performance of humans and machines.

\subsection{Comparing Human and Machine Performance}
\label{depth:sec:compare-human-vs-machine}

Table~\ref{depth:tbl:classes-vs-accuracy} shows \countDepthExperiments{} skin condition classification experiments selected from \countDepthStudies{} studies, spanning both clinical and computing research. The works in this table were selected based on the following criteria: 1) they compared humans and machines over the same dataset, or 2) they reported human and machine performance separately on a multi-class (greater than two) dataset. Experiments where the \emph{predictions} (Eq.~\ref{depth:eq:predictions}) of a model could not be inferred were omitted. This primarily occurred when only the AUROC scores were reported.

As not all studies report accuracy, accuracy was inferred given the other reported metrics. 
Occasionally, the exact sensitivity and specificity were not given, and these were estimated from the reported graphs. For studies that had predictions made by multiple humans, the accuracy was computed using the average human performance.

\begin{figure}[ht]
    \centering
    \includegraphics[width=0.48\textwidth]{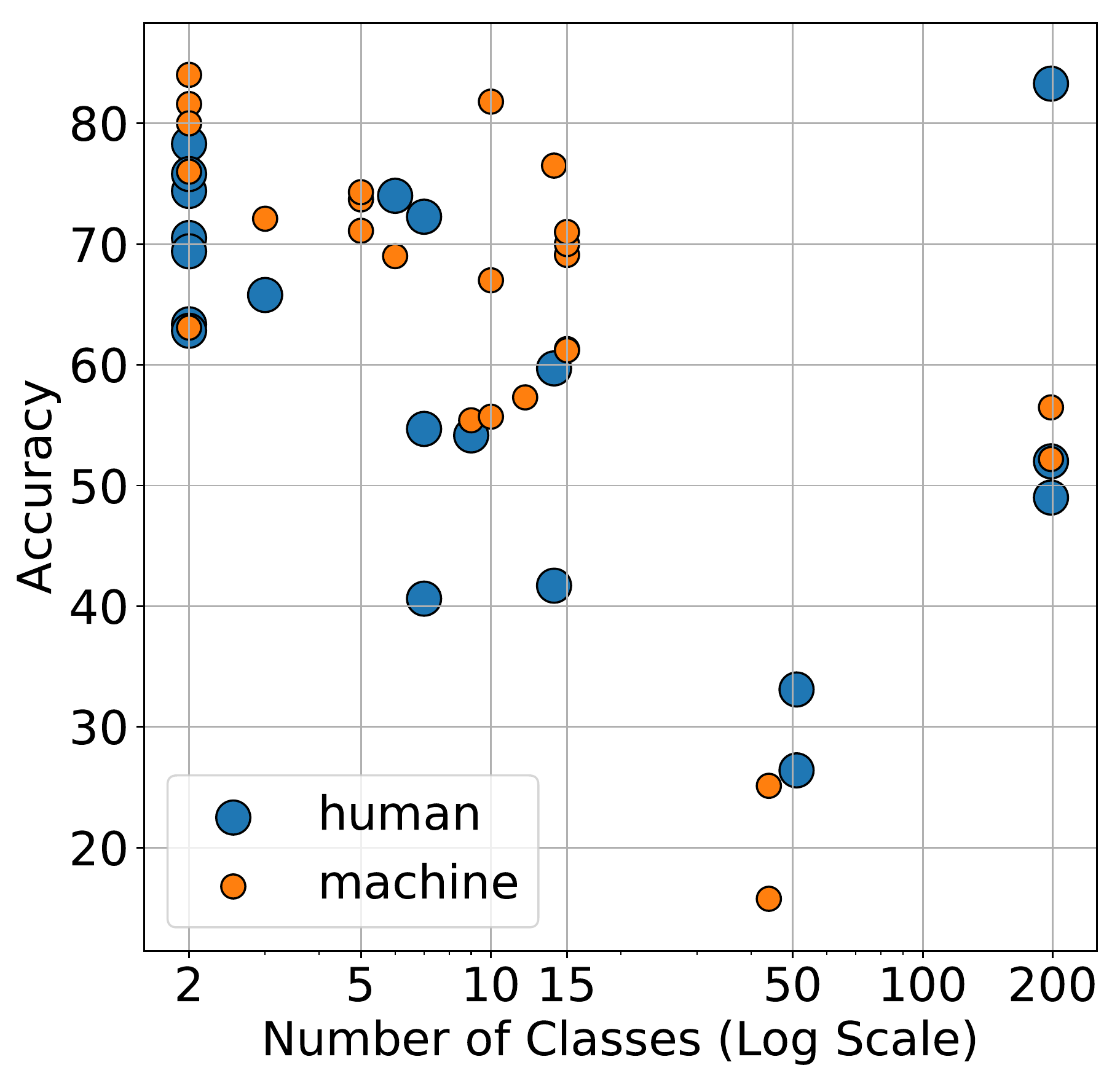}
    \caption{Skin disease classes versus reported model accuracy. Each coloured dot represents a experiment from Table~\ref{depth:tbl:classes-vs-accuracy}, where the diagnosis was made by either a human or machine.
    }
    \label{depth:fig:classes-vs-accuracy}
\end{figure}

Table~\ref{depth:tbl:classes-vs-accuracy} reports the dataset and the number of images used to evaluate, the input modality, whether human or machines did the diagnosis, the number of classes, and the accuracy over the entire test set. Fig.~\ref{depth:fig:classes-vs-accuracy} plots the number of classes versus the reported accuracy, separated based on machine and human skin disease classification performance. A general trend is observed, where as the number of classes increases, the accuracy decreases.

\begin{figure}[ht]
    \centering
    \includegraphics[width=0.48\textwidth]{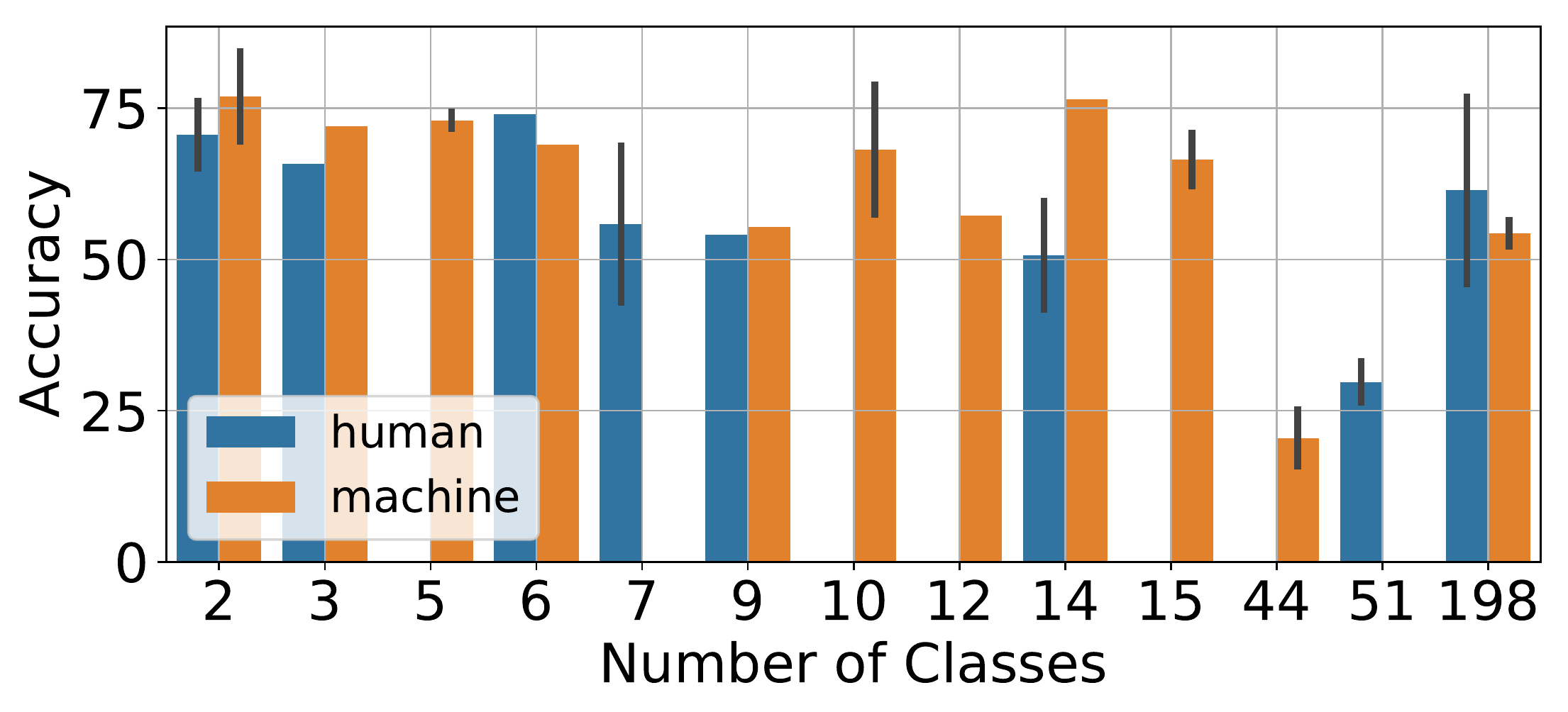}
    \caption{Averaged accuracy of the experiments in Table~\ref{depth:tbl:classes-vs-accuracy}, grouped by the number of classes. On average, similar performance of both human and machines is reported.}
    \label{depth:fig:avg-acc-by-class}
\end{figure}

\begin{figure}[ht]
    \centering
    \includegraphics[width=0.48\textwidth]{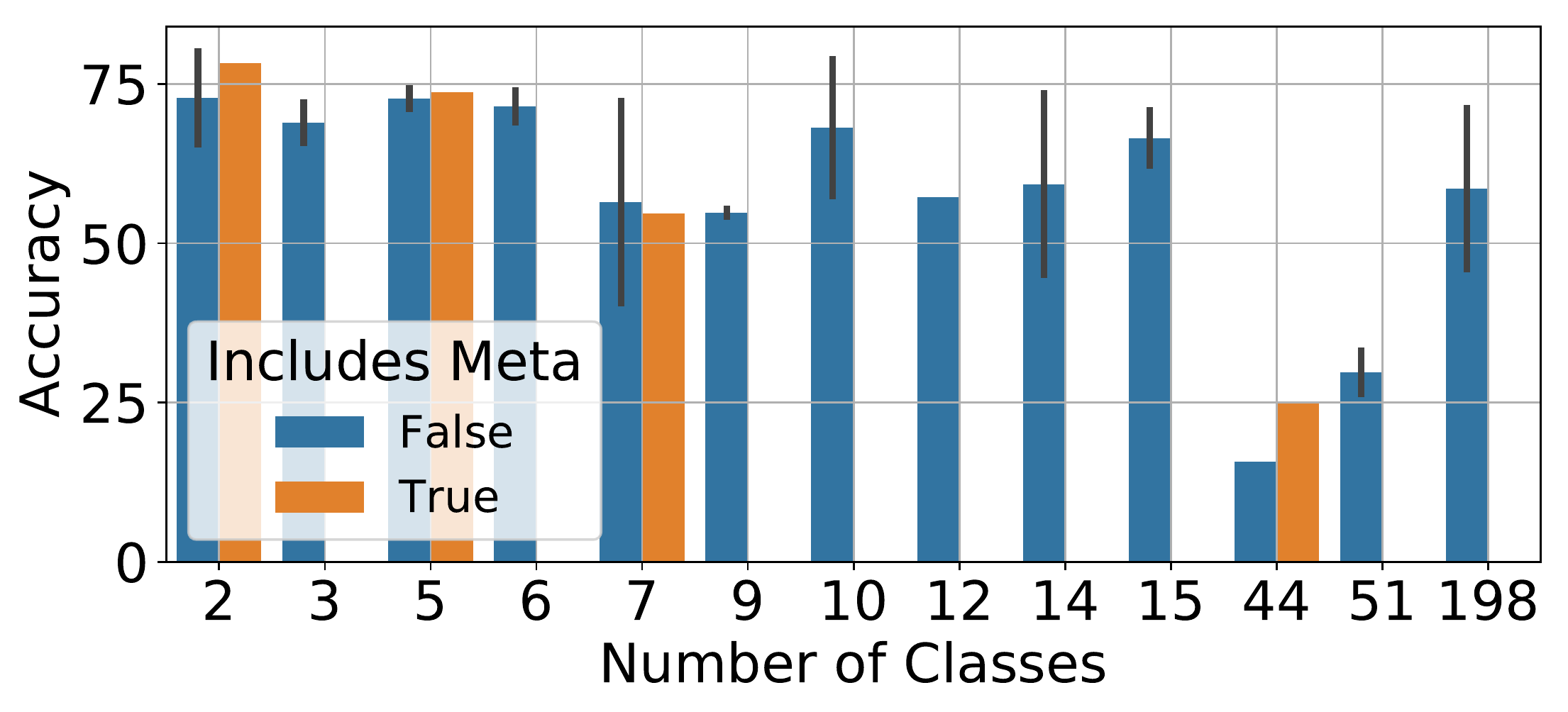}
    \caption{Averaged accuracy of the experiments in Table~\ref{depth:tbl:classes-vs-accuracy}, grouped by the number of classes. Similar accuracy is reported when additional meta data (\ie patient history) is included in the diagnosis.}
    \label{depth:fig:meta}
\end{figure}

Similar accuracy is found when averaged across studies for both humans and machines (Fig.~\ref{depth:fig:avg-acc-by-class}). As well, Fig.~\ref{depth:fig:meta} highlights that the inclusion of patient history (\eg questionnaire, age, sex) yields small changes to accuracy, with the exception of one non-deep learning study that included 37 user supplied answers~\cite{Razeghi2014}.

\begin{table*}[ht]
\centering
\caption{Selected skin disease classification approaches and diagnostic performance. \emph{N.Images} indicates the number of images in the dataset. \emph{N.Test} indicates the number of images used to test (includes cross-validation). \emph{Derm.}, \emph{Clinic.}, and \emph{Meta.} indicate dermoscopy images, clinical images, and non-image patient history (meta-data), respectively, where a star (*) indicates \emph{in vivo} data. \emph{H.vs.M} indicates if the diagnosis was made by a human or machine. \emph{Acc.} indicates diagnostic accuracy.}
\small
\begin{tabular}{@{~}c c c c @{~}c @{~}c@{~}c@{~}@{~}c@{~}c c c@{~}}
\hline
                      &  Year &   Dataset & N.Images &   N.Test &        Derm. &      Clinic. &         Meta &   H.vs.M &  Classes &   Acc. \\
\hline
    \cite{Ferris2015} &  2015 &  Internal &        - &       65 &   \checkmark &           &           &    human &        2 &  63.35 \\
    \cite{Ferris2015} &  2015 &  Internal &      273 &       65 &   \checkmark &           &           &  machine &        2 &  63.08 \\
   \cite{Codella2017} &  2017 &  ISIC-100 &        - &      100 &   \checkmark &           &           &    human &        2 &  70.50 \\
   \cite{Codella2017} &  2017 &  ISIC-100 &     1000 &      100 &   \checkmark &           &           &  machine &        2 &  76.00 \\
  \cite{Haenssle2018} &  2018 &  Internal &        - &      100 &   \checkmark &           &           &    human &        2 &  74.40 \\
  \cite{Haenssle2018} &  2018 &  Internal &        - &      100 &   \checkmark &           &   \checkmark &    human &        2 &  78.30 \\
  \cite{Haenssle2018} &  2018 &  Internal &       &      100 &   \checkmark &           &           &  machine &        2 &  81.60 \\
       \cite{Han2018} &  2018 &      Asan &        - &     1133 &           &   \checkmark &           &    human &        2 &  75.80 \\
       \cite{Han2018} &  2018 &      Asan &   49,567 &     1133 &           &   \checkmark &           &  machine &        2 &  80.00 \\
  \cite{Brinker2019a} &  2019 &  ISIC-100 &    13737 &      100 &   \checkmark &           &           &  machine &        2 &  84.02 \\
   \cite{Brinker2019} &  2019 &  ISIC-100 &        - &      100 &   \checkmark &           &           &    human &        2 &  62.82 \\
   \cite{Brinker2019} &  2019 &  MED-NODE &        - &      100 &           &   \checkmark &           &    human &        2 &  69.40 \\
    \cite{Esteva2017} &  2017 &  Stanford &        - &      180 &   \checkmark &   \checkmark &           &    human &        3 &  65.78 \\
    \cite{Esteva2017} &  2017 &  Stanford &  127,463 &  127,463 &   \checkmark &   \checkmark &           &  machine &        3 &  72.10 \\
 \cite{Ballerini2013} &  2013 &  Dermofit &      960 &      960 &           &   \checkmark &           &  machine &        5 &  74.30 \\
  \cite{Kawahara2018} &  2018 &     Atlas &     2018 &      395 &   \checkmark &   \checkmark &           &  machine &        5 &  71.10 \\
  \cite{Kawahara2018} &  2018 &     Atlas &     2018 &      395 &   \checkmark &   \checkmark &   \checkmark &  machine &        5 &  73.70 \\
  \cite{Tschandl2017} &  2017 &  Internal &      348 &       50 &   \checkmark &           &           &    human &        6 &  74.00 \\
  \cite{Tschandl2017} &  2017 &  Internal &      348 &       50 &   \checkmark &           &           &  machine &        6 &  69.00 \\
     \cite{Carli2002} &  2002 &  Internal &        - &      256 &           &  \checkmark* &  \checkmark* &    human &        7 &  40.62 \\
     \cite{Carli2002} &  2002 &  Internal &        - &      256 &   \checkmark &           &   \checkmark &    human &        7 &  54.69 \\
     \cite{Carli2002} &  2002 &  Internal &        - &      256 &  \checkmark* &  \checkmark* &  \checkmark* &    human &        7 &  72.27 \\
    \cite{Esteva2017} &  2017 &  Stanford &        - &      180 &   \checkmark &   \checkmark &           &    human &        9 &  54.15 \\
    \cite{Esteva2017} &  2017 &  Stanford &  127,463 &  127,463 &   \checkmark &   \checkmark &           &  machine &        9 &  55.40 \\
       \cite{Leo2015} &  2015 &  Dermofit &     1300 &     1300 &           &   \checkmark &           &  machine &       10 &  67.00 \\
  \cite{Kawahara2016} &  2016 &  Dermofit &     1300 &     1300 &           &   \checkmark &           &  machine &       10 &  81.80 \\
      \cite{Han2018c} &  2018 &  Dermofit &   20,689 &     1300 &           &   \checkmark &           &  machine &       10 &  55.70 \\
      \cite{Han2018c} &  2018 &      Asan &   19,389 &    1,276 &           &   \checkmark &           &  machine &       12 &  57.30 \\
  \cite{Fujisawa2019} &  2019 &  Internal &        - &     1260 &           &   \checkmark &           &    human &       14 &  41.70 \\
  \cite{Fujisawa2019} &  2019 &  Internal &        - &     1820 &           &   \checkmark &           &    human &       14 &  59.70 \\
  \cite{Fujisawa2019} &  2019 &  Internal &     6009 &     1142 &           &   \checkmark &           &  machine &       14 &  76.50 \\
  \cite{Demyanov2017} &  2017 &   MoleMap &   40,173 &     1776 &   \checkmark &   \checkmark &           &  machine &       15 &  69.10 \\
       \cite{Ge2017a} &  2017 &   MoleMap &    26584 &     7975 &           &   \checkmark &           &  machine &       15 &  61.20 \\
       \cite{Ge2017a} &  2017 &   MoleMap &    26584 &     7975 &   \checkmark &           &           &  machine &       15 &  61.30 \\
       \cite{Ge2017a} &  2017 &   MoleMap &    26584 &     7975 &   \checkmark &   \checkmark &           &  machine &       15 &  70.00 \\
        \cite{Ge2017} &  2017 &   MoleMap &   32,194 &    8,012 &   \checkmark &   \checkmark &           &  machine &       15 &  71.00 \\
   \cite{Razeghi2014} &  2014 &    dermis &     2309 &     1429 &           &   \checkmark &           &  machine &       44 &  15.76 \\
   \cite{Razeghi2014} &  2014 &    dermis &     2309 &     1429 &           &   \checkmark &   \checkmark &  machine &       44 &  25.12 \\
      \cite{Sinz2017} &  2017 &  Internal &        - &     2072 &           &   \checkmark &           &    human &       51 &  26.40 \\
      \cite{Sinz2017} &  2017 &  Internal &        - &     2072 &   \checkmark &           &           &    human &       51 &  33.10 \\
       \cite{Sun2016} &  2016 &    SD-198 &    6,584 &     3292 &           &   \checkmark &           &  machine &      198 &  52.19 \\
      \cite{Yang2018} &  2018 &    SD-198 &        - &       &           &   \checkmark &           &    human &      198 &  49.00 \\
      \cite{Yang2018} &  2018 &    SD-198 &        - &       &           &   \checkmark &           &    human &      198 &  52.00 \\
      \cite{Yang2018} &  2018 &    SD-198 &        - &       &           &   \checkmark &           &    human &      198 &  83.29 \\
      \cite{Yang2018} &  2018 &    SD-198 &     6584 &     3292 &           &   \checkmark &           &  machine &      198 &  56.47 \\
\hline
\end{tabular}
\label{depth:tbl:classes-vs-accuracy}
\end{table*}

\subsection{Limitations and Sources of Errors in Image-Based Diagnosis}
When developing an image-based classification system, there are several limitations and sources of potential errors. The ``ground truth'' disease labels may have errors, even when confirmed via histopathology. Monheit \etal~\cite{Monheit2011} found that due to conflicts in the expert histopathology diagnoses, 8.8\% of lesions required more than two histopathological evaluations before reaching a final diagnosis. Elmore \etal~\cite{Elmore2017} collected 240 biopsy cases and used the consensus of three human experts to label each case into one of five categories that indicated a progressively increasing melanoma risk. These consensus labels were compared to diagnoses given by 187 pathologists, and the authors found that the three diagnoses categories spanning ``moderately dysplastic nevi to early stage invasive melanoma were neither reproducible nor accurate''~\cite{Elmore2017}. 

Patients may be limited by their ability to capture high quality images. Weingast \etal~\cite{Weingast2013} had patients attempt to acquire an image of their own lesion using a mobile phone camera. However, 81\% of patients required assistance in acquiring images, partly due hard-to-reach lesions, and challenges in focusing and choosing an appropriate field-of-view. Even with assistance, 39\% of the cases were reported to have insufficient information to make a diagnosis via teledermatology (but could be diagnosed face-to-face), indicating significant challenges in acquiring quality images.

Images may contain insufficient or misleading information easily resolved during a face-to-face examination. In Sec.~\ref{depth:sec:derm-static}, the reported differences in human diagnosis performance during teledermatology suggests that diagnosing via static images may be significantly more challenging than diagnosis during face-to-face consultations. A further example is given by, Hogan \etal~\cite{Hogan2015} who documented a patient supplied image that appeared to contain serious complications, but on a face-to-face inspection revealed a crust covering a well-healing wound. Thus, claims that machines have reached human-level diagnostic ability should be considered in the context of static images. 

The role non-visual information (\eg patient history, questionnaire data) takes in the diagnostic procedure and what information should be gathered is not clear. Machine diagnoses systems that do utilize non-visual patient data report a mixed impact to performance, ranging from minimal~\cite{Kawahara2018} to substantial improvements~\cite{Razeghi2014}. Experienced dermatologists exhibit minimal improvements to diagnosis when given patient history (age, sex, body location site) in addition to an image, but those with less experience show a greater improvement with access to patient history~\cite{Haenssle2018}. 
Acquiring this data outside of face-to-face consultations may also be challenging. Weingast \etal~\cite{Weingast2013} reported that most patients over 60 years needed assistance completing a computer questionnaire. 

Another consideration is how transferable across datasets and populations the models are. Han \etal~\cite{Han2018c} report an accuracy of 55.7\% over the 10-classes of Dermofit. This is significantly lower than other reported works that train and test only over Dermofit (\eg 81\%~\cite{Kawahara2016}). As Han \etal~\cite{Han2018c}'s model was trained on an Asian population and tested on a European population, this drop in accuracy may be due to the differences in how skin diseases manifest across populations, or signify a lack of transferability in learned features across datasets due to image acquisition protocols, or both. 

These sources of errors and limitations are potentially compounding, where ground truth errors in training, may compound with low quality acquired images, and a lack of model transferability across populations.

\section{Conclusions}
While there are still significant challenges in skin disease diagnosis, in 2017 dermatologists from a variety of institutions wrote the following statement~\cite{Lim2017}: 
\begin{quote}
``With the physician workforce projected to remain relatively flat, the specific ratio of dermatologists to population will decrease over time, especially in rural areas. These projections indicate a current and future challenge to ensure patient access to appropriate dermatologic care.''
\end{quote}
Automated analysis of skin conditions has the potential to alleviate the diagnostic requirements of dermatologists, making this a field worthy of investigation.

Given that recent studies report comparable accuracy performance when comparing dermatologists and machines, and considering the reported performance across independent teledermatological and machine studies, it is reasonable to conclude that machine accuracy is nearing the performance of human dermatologists in a teledermatological scenario. However, given the differences in performance when dermatologists diagnose via teledermatology~\cite{Weingast2013}, machine classification accuracy may be significantly lower than a face-to-face consultation with a dermatologist. As the diagnostic performance of general practitioners is reported to be twice as low as dermatologists~\cite{Sellheyer2005}, machine classification of skin diseases may have increased utility among general practitioners, who are often the first clinicians to examine dermatological disorders. We highlight that when humans classify among the 1000-classes of the natural images within ImageNet, the reported top-5 error (considers a match in any of the top-5 predictions to be correct) is 5.1\%-12\%~\cite{Russakovsky2014}. While not directly comparable due to the different number of possible classes considered, the relatively low accuracy for humans classifying skin diseases commonly reported in Table~\ref{depth:tbl:classes-vs-accuracy} indicates the challenges of classifying skin diseases from images. Finally, we note that diagnostic accuracy, which is focused on in this review, gives us a limited understanding of performance, and does not consider the severity of misdiagnosing certain conditions.

\bibliographystyle{IEEEtran}
\bibliography{IEEEabrv,main}

\end{document}